\documentclass[10pt,twocolumn,letterpaper]{article}

\usepackage{wacv}
\usepackage{times}
\usepackage{epsfig}
\usepackage{graphicx}
\usepackage{amsmath}
\usepackage{amssymb}
\usepackage{hyperref}

\wacvfinalcopy 

\def\wacvPaperID{443} 

\ifwacvfinal\fi
\newcommand{\Tref}[1]{Table~\ref{#1}}
\newcommand{\Eref}[1]{Equation~(\ref{#1})}
\newcommand{\Fref}[1]{Figure~\ref{#1}}

\graphicspath{ {images/} }
\begin{document}

\title{Gated Context Aggregation Network for Image Dehazing and Deraining}

\author{Dongdong Chen$^1$, \quad Mingming He$^2$, \quad  Qingnan Fan$^3$,\quad  Jing Liao$^4$\\ Liheng Zhang$^5$, \quad Dongdong Hou$^1$, \quad Lu Yuan$^6$, \quad Gang Hua$^6$\\
	$^{1}$University of Science and Technology of China, \qquad $^{2}$ Hong Kong University of Science and Technology \\
\qquad $^{3}$Shandong University,  \qquad $^{4}$City University of Hong Kong \\
\qquad $^{5}$University of Central Florida, \qquad $^{6}$Microsoft Cloud and AI \\
	{\tt\small cd722522@mail.ustc.edu.cn}, \quad{\tt\small hmm.lillian@gmail.com}, \quad{\tt\small fqnchina@gmail.com}, \\ \quad{\tt\small liaojing8871@gmail.com}, \quad{\tt\small lihengzhang1993@knights.ucf.edu}, \\ \quad{\tt\small houdd@mail.ustc.edu.cn}, \quad{\tt\small \{luyuan, ganghua\}@microsoft.com} }

\maketitle
\ifwacvfinal\thispagestyle{empty}\fi

\begin{abstract}
Image dehazing aims to recover the uncorrupted content from a hazy image. Instead of leveraging traditional low-level or handcrafted image priors as the restoration constraints, e.g., dark channels and increased contrast, we propose an end-to-end gated context aggregation network to directly restore the final haze-free image. In this network, we adopt the latest smoothed dilation technique to help remove the gridding artifacts caused by the widely-used dilated convolution with negligible extra parameters, and leverage a gated sub-network to fuse the features from different levels. Extensive experiments demonstrate that our method can surpass previous state-of-the-art methods by a large margin both quantitatively and qualitatively. In addition, to demonstrate the generality of the proposed method, we further apply it to the image deraining task, which also achieves the state-of-the-art performance.  Code has been made available at \url{https://github.com/cddlyf/GCANet}.
\end{abstract}

\section{Introduction}
Due to the existence of turbid medium (e.g., dusk, smoke, and other particles) in the atmosphere, images taken in such atmospheric phenomena are subject to visible quality degradation, such as contrast and saturation loss. Taking these degraded images as input, many vision-based systems, originally designed with the assumption of clean capture environments, may be easily troubled with drastic performance decrease. Given that, image dehazing has been extensively studied to restore the clean image from the corrupted input, to serve as the preprocessing step of the aforementioned systems. 

In this literature, the hazing processing is often represented with the physical corruption model:
\begin{equation} \label{eq:dehaze_model}
\mathbf{I}(x) = \mathbf{J}(x)t(x) + \mathbf{A}(1-t(x))
\end{equation}
where $\mathbf{I}(x)$ and $\mathbf{J}(x)$ are the degraded hazy image and the target haze-free scene radiance respectively. $\mathbf{A}$ is the global atmospheric light, and $t(x)$ is the medium transmission map, which is dependent on the unknown depth information. Most previous dehazing methods first estimate the transmission map $t(x)$ or the atmospheric light $\mathbf{A}$,  then try to recover the final clean image $\mathbf{J}(x)$. But the first step is a very challenging problem because both the transmission map $t(x)$ and the atmospheric light $\mathbf{A}$ are often unknown in the real scenarios. 

To compensate for the lost information during the corruption procedure, many traditional methods \cite{berman2016non,hautiere2007towards,he2011single,meng2013efficient,pei2012nighttime,zhu2015fast} leverage some image priors and visual cues to estimate the transmission maps and atmospheric light. For example, \cite{hautiere2007towards} maximizes the local contrast of the target image by using the prior that the contrast of degraded images is often drastically decreased. \cite{he2011single} proposes the dark channel prior based on the assumption that image patches of outdoor haze free images often have low-intensity values. \cite{berman2016non} relies on the assumption that haze-free image colors are well approximated by a few hundred distinct colors and proposes a non-local prior-based dehazing algorithm. However, these priors do not always hold, so they may not work well in certain real cases.

With the latest advances of deep learning, many CNN-based methods \cite{ancuti2013single,cai2016dehazenet,ren2016single,li2017aod,ren2018gated,zhang2018densely} are proposed by leveraging a large scale training datasets. Compared to traditional methods as described above, CNN-based methods attempt to directly regress the intermediate transmission map or the final clean image, and achieve superior performance and robustness. \cite{cai2016dehazenet} presents an end-to-end network to estimate the intermediate transmission map. \cite{li2017aod} reformulates the atmospheric scattering model to predict the final clean image through a light-weight CNN. \cite{ren2018gated} creates three different derived input images from the original hazy image and fuses the dehazed results out of these derived inputs. \cite{zhang2018densely} incorporates the physical model in \Eref{eq:dehaze_model} into the network design and uses two sub-networks to regress the transmission map and atmospheric light respectively.

In this paper, we propose a new end-to-end gated context aggregation network (denoted as "\textbf{GCANet}") for image dehazing. Since dilated convolution is widely used to aggregate context information for its effectiveness without sacrificing the spatial resolution \cite{yu2015multi,li2018recurrent,wang2018understanding,hamaguchi2018effective,fan2018decouple}, we also adopt it to help obtain more accurate restoration results by covering more neighbor pixels. However, the original dilated convolution will produce so-called "gridding artifacts" \cite{wang2018understanding,hamaguchi2018effective}, because adjacent units in the output are computed from completely separate sets in the input when the dilation rate is larger than one. Recently, \cite{wang2018smoothed} analyzes the dilation convolution in a compositional way and proposes to smooth the dilated convolution, which can greatly reduce this gridding artifacts. Hence, we also incorporate this idea in our context aggregation network. As demonstrated in \cite{zhang2018densely,Lin_2017_CVPR}, fusing  different levels of features is often beneficial for both low-level and high-level tasks. Inspired by it, we further propose a gated sub-network to determine the importance of different levels and fuse them based on their corresponding importance weights. \cite{ren2018gated} also uses a gated fusion module in their network, but they directly fuse the dehazing results of different derived input images rather than the intermediate features. 

To validate the effectiveness of the proposed \textbf{GCANet}, we compare it with previous state-of-the-art methods on the recent dehazing benchmark dataset RESIDE \cite{li2017reside}. Experiments demonstrate that our \textbf{GCANet} outperforms all the previous methods both qualitatively and quantitatively by a large margin. Furthermore, we conduct comprehensive ablation studies to understand the importance of each component. To show the generality of the proposed \textbf{GCANet}, we have also applied it to the image deraining task, which can also obtain superior performance over previous state-of-the-art image deraining methods.

To summarize, our contributions are three-fold as below:
\begin{itemize}
\item We propose a new end-to-end gated context aggregation network \textbf{GCANet} for image dehazing, in which the smoothed dilated convolution is used to avoid the gridding artifacts and a gated subnetwork is applied to fuse the features of different levels.
\item Experiments show that \textbf{GCANet} can obtain much better performance than all the previous state-of-the-art image dehazing methods both qualitatively and quantitatively. We also provide comprehensive ablation studies to validate the importance and necessity of each component.
\item We further apply our proposed \textbf{GCANet} to the image deraining task, which also outperforms previous state-of-the-art image deraining methods and demonstrates its generality.
\end{itemize}

The remainder of the paper is organized as follows. We will first summarize related work in Section 2, then give our main technical details in Section 3. Finally, we will provide comprehensive experiments results and ablation studies in Section 4 and conclude in Section 6.

\section{Related Work}
Single image dehazing is the inverse recovery procedure of the physical corruption procedure defined in \Eref{eq:dehaze_model}, which is a highly ill-posed problem because of the unknown transmission map and global atmospheric light. In the previous several decades, many different image dehazing methods are proposed to tackle this challenging problem, which can be roughly divided into traditional prior-based methods and modern learning-based methods. The most significant difference between these two types is that the image priors are handcrafted in the former type but are learned automatically in the latter type.

In the traditional prior-based methods, many different image statistics priors are leveraged as extra constraints to compensate for the information loss during the corruption procedure. For example, \cite{fattal2008single} propose a physically grounded method by estimating the albedo of the scene. \cite{he2011single,xie2010improved,xu2012fast} discover and improve the effective dark channel prior to calculate the intermediate transmission map more reliably. \cite{tan2008visibility} use Markov Random Field to maximize the local contrast of an image by assuming that the local contrast of a clear image is higher than that of a hazy image.  Based on the observation that small image patches typically exhibit a one-dimensional distribution in the RGB color space, \cite{fattal2014dehazing} recently propose a color-line method for image dehazing and \cite{berman2016non} propose a non-local path prior to characterize the clean images. These dedicatedly handcrafted priors , however, hold for some cases, but they are not always robust to handle all the cases.

Recently, learning-based methods are proposed for image dehazing by leveraging the large-scale datasets and the powerful parallelism of GPU. In these type of methods, the image priors are automatically learned from the training dataset by the neural network and saved in the network weights. Their main differences typically lie in the learning targets and the detailed network structures. \cite{cai2016dehazenet,ren2016single} propose an end-to-end CNN network and multi-scale network respectively to predict the intermediate transmission maps. However, inaccuracies in the estimation of the transmission map always lead to low-quality dehazed results. \cite{li2017aod} encode the transmission map and the atmospheric light into one variable, and then use a lightweight network to predict it. \cite{zhang2018densely} design two different sub-networks for the prediction of the transmission map and the atmospheric light by following the physical model defined in \Eref{eq:dehaze_model}. We propose an end-to-end gated context aggregation network for image dehazing but different from these methods, our proposed \textbf{GCANet} is designed to directly regress the residue between the hazy image and the target clean image. Moreover, our network structure definitely distinguish from the previous ones, which is quite lightweight but can achieve much better results than all the previous  methods.

\begin{figure*}[ht]
	\centering \includegraphics[width=0.975\textwidth]{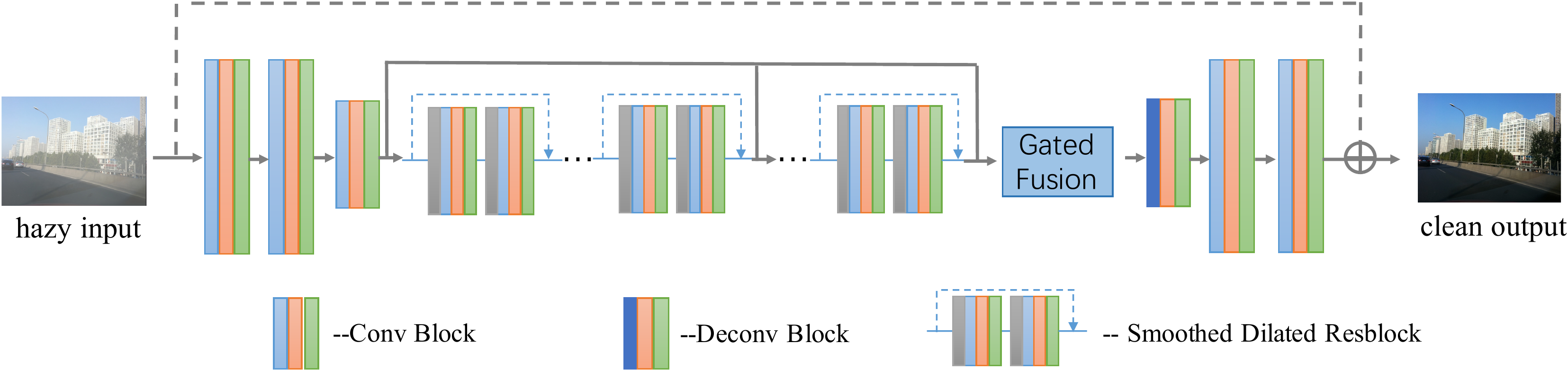}
\caption{The overall network structure of the proposed \textbf{GCANet}, which follows a basic auto-encoder structure. It consists of three convolution blocks as the encoder part, and one deconvolution block and two convolution blocks as the decoder part. Several smoothed dilated resblocks are inserted between them to aggregate context information without gridding artifacts. To fuse the features from different levels, an extra gate fusion sub-network is leveraged. During the runtime, the \textbf{GCANet} will predict the residue between the target clean image and the hazy input image in an end-to-end way. }
	\label{fg:arch}
\end{figure*}

\section{Method}
In this section, we will introduce the architecture of the proposed gated context aggregation network \textbf{GCANet}. As shown in \Fref{fg:arch}, given a hazy input image, we first encode it into feature maps by the encoder part, then enhance them by aggregating more context information and fusing the features of different levels  without downsampling. Specifically,  the smoothed dilated convolution and an extra gate sub-network are leveraged. The enhanced feature maps will be finally decoded back to the original image space to get the target haze residue. By adding it onto the input hazy image, we will get the final haze free image. 

\paragraph{Smoothed Dilated Convolution}Modern image classification networks \cite{krizhevsky2012imagenet,szegedy2015going,he2016deep} often integrate multi-scale contextual information via successive pooling and subsampling layers that reduce resolution until a global prediction is obtained. However, for dense prediction tasks like segmentation, the contradiction is the required multi-scale  contextual reasoning and the lost spatial resolution information during downsampling. To solve this problem, \cite{yu2015multi} proposes a new dilated convolutional layer, which supports exponential expansion of the receptive field without loss of resolution or coverage. In the one-dimension case, given a 1-D input $f$, the output of the regular convolutional layer $w$ with kernel size $k$ is:
\begin{equation}
(f \otimes w)(i) = \sum_{j=1}^{k} f[i+j]w[j]  
\end{equation}
where one output point cover total $k$ input points, so the receptive field is $k$. But for the dilated convolution, it can be viewed as "convolution with a dilated filter", which can be represented as:
\begin{equation}
(f \otimes_r w)(i) = \sum_{j=1}^{k} f[i+r*j]w[j]  
\end{equation}
where $r$ is the dilation rate, and the dilated convolution will degenerate to regular convolution when $r=1$. To understand the dilated convolution in an intuitive way, we can view it as inserting $r-1$ zeros between two adjacent weights of $w$. In this way, the dilated convolution can increase the original receptive field from $k$ to $r*(k-1)+1$ without reducing the resolution.

\begin{figure}[t]
	\centering \includegraphics[width=0.45\textwidth]{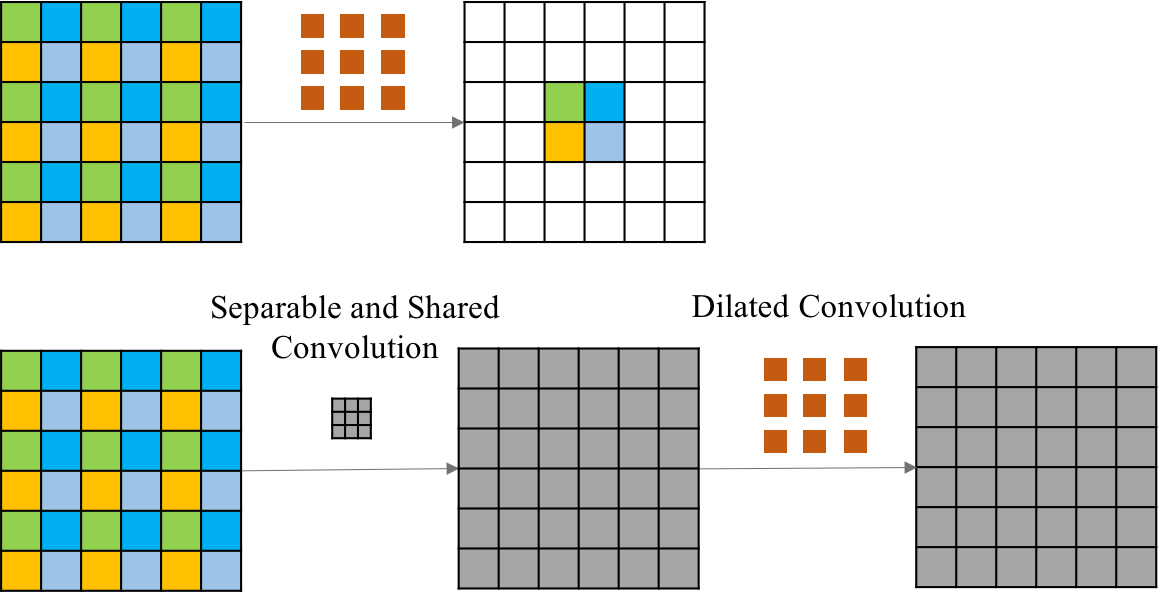}
\caption{The illustration of gridding artifacts of dilated convolution and the proposed smoothed dilated convolution in  \cite{wang2018smoothed}: the four different points in next layer $i$ are indicated by different colors, it can be seen that they are related to completely different sets of units of previous layer, which will potential cause  the gridding artifacts. By contrast,  the smoothed dilated convolution, which adds the dependency among the input units with an extra separable and shared convolutional layer before the dilated convolution.}
	\label{fg:dilated_grid}
\end{figure}

Despite of the effectiveness of the dilated convolution, it will produce the so-called gridding artifacts, which is also noticed in previous papers \cite{wang2018understanding,hamaguchi2018effective}. To understand this issue more clearly, a very recent work \cite{wang2018smoothed} analyzes the dilated convolution in a compositional way. The illustration of gridding artifacts is shown in \Fref{fg:dilated_grid}, where the case of one dilated convolutional layers with $r=2, k=3$ is analyzed. Considering the four neighbor pixels of the next layer, they and their dependent units in the previous layer are marked with four different colors respectively. We can easily find that these four neighor pixels are related to totally different sets of previous units in the previous layer. In other words, there is no dependency among the input units or the output units in the dilated convolution. This is why it will potentially cause the inconsistencies, i.e. gridding artifacts.

To alleviate it, \cite{wang2018smoothed} proposes to add interaction among the input units before dilated convolution or output units after dilated convolution by adding an extra convolutional layer of kernel size $(2r-1)$. In this paper, we choose to add the dependency of input units by default. Need to note that, \cite{wang2018smoothed} adopts a separable and shared convolution as the extra convolutional layer rather than the vanilla one. ``Separable'' means the separable convolution idea from \cite{chollet2017xception}, while ``shared'' means the convolution weights are shared for all the channels. In this way, this special convolutional layer has a constant parameter size $(2r-1)^2$, which is independent of the feature channel number. \Fref{fg:dilated_grid} is one illustration of smoothed dilated convolution.

\paragraph{Gated Fusion Sub-network} As shown in \cite{Lin_2017_CVPR,zhang2018densely}, fusing the features from different levels is often beneficial both for low-level and high-level tasks. To implement this idea, \cite{Lin_2017_CVPR} uses the feature pyramids to fuse high-level semantic feature maps at all scales, and \cite{zhang2018densely} leverages the densely connected networks. In this paper, we adopt a different way by incorporation of an extra gated fusion sub-network $\mathcal{G}$. Specifically, we first extract the feature maps from different levels $F_l, F_m, F_h$, and feed them into the gated fusion sub-network. The output of the gated fusion sub-network are three different importance weights $(\mathcal{M}_l, \mathcal{M}_m, \mathcal{M}_h)$, which correspond to each feature level respectively. Finally, these three  features maps $F_l, F_m, F_h$ from different levels are linearly combined with the regressed importance weights.

\begin{equation}
\begin{aligned}
(\mathcal{M}_l, \mathcal{M}_m, \mathcal{M}_h) = \mathcal{G}(F_l, F_m, F_h) 
\\
F_o = \mathcal{M}_l * F_l + \mathcal{M}_m * F_m + \mathcal{M}_h * F_h
\end{aligned}
\end{equation}

The combined feature map $F_o$ will be further fed into the decoder to get the target haze residue. In this paper, our gated fusion sub-network consists of only one convolutional layer with kernel size 3x3, whose input is the concatenation of $F_l, F_m, F_h$ and output channel number is 3.

\paragraph{Network Structure}Following the similar network design principle in \cite{johnson2016perceptual,fan2017generic,fan2018decouple}, our overall network structure are also designed as a simple auto-encoder, where seven residual blocks are inserted between the encoder and decoder to enhance its learning capacity. Specifically, three convolutional layers are first used to encode the input hazy image into the feature maps as the encoder part, where only the last convolutional layer downsamples the feature maps by 1/2 once. Symmetrically, one deconvolutional layer with stride 1/2 is used to upsample the feature map to the original resolution in the decoder part, then the following two convolutional layers convert the feature maps back to the image space to get the final target haze residue. For the intermediate residual blocks, we call them ``Smoothed Dilated Resblock" , because we have replaced all the original regular convolutional layers with the aforementioned smoothed dilated convolutional layers. The dilation rates of these seven residual blocks are setted as  $(2,2,2,4,4,4,1)$ respectively. To obtain a good tradeoff between the performance and runtime, we set the channel number of all the intermediate convolutional layers as 64. Note that except for the last convolutional layer and every extra separable and shared convolutional layer in the smoothed dilated convolution layer, we put an instance normalization layer \cite{ulyanovinstance} and ReLU layer after each convolutional layer. In the experiment part, we will show instance normalization is more suitable than batch normalization for the image dehazing task.

As demonstrated in \cite{fan2017generic,fan2018decouple}, besides the input image, pre-calculating the edge of the input image and feeding them into the network as the auxiliary information is  very helpful to the network learning. Hence, by default, we also adopt this simple idea and concatenate the pre-calculated edge with the input hazy image along the channel dimension as the final inputs of \textbf{GCANet}.

\paragraph{Loss Function}
In previous learning-based image dehazing methods \cite{cai2016dehazenet,ren2016single,li2017aod,li2018single,zhang2018densely,zhang2018multi}, the simple Mean Square Error loss is  adopted. Following the same strategy, we also use this simple loss by default. But different from these methods, our learning target is the residue between the  haze free image and the input hazy one:
\begin{equation}\label{eq:loss}
\begin{aligned}
r &= J - I
\\
\hat{r} &= GCANet(I)
\\
\mathcal{L} &= \lVert \hat{r}  - r \rVert^2
\end{aligned}
\end{equation}
where $r$ and $\hat{r}$ are the ground truth and predicted haze residue respectively. During runtime, we will add $\hat{r}$ onto the input hazy image to get the final predicted haze free image. Need to emphasize that designing better loss function is not the focus of this paper, but our proposed \textbf{GCANet} should be able to generalize to better designed losses. For example, \cite{li2018single,zhang2018densely,zhang2018multi} find the perceptual loss \cite{johnson2016perceptual} and GAN loss can improve the final dehazing results. However, even only with the above simple loss, our method can still achieve the state-of-the-art performance.

\section{Experiments}
\paragraph{Implementation Details} For experiments, we first validate the effectiveness of the proposed \textbf{GCANet} on the image dehazing task, then demonstrate its generality by further applying it to image deraining task. To train these two tasks, we all directly adopt the available benchmark datasets both for training and evaluation. For each task, we compare our method with many previous state-of-the-art methods.  Without losing generality, we use almost the same training strategy for these two tasks. By default, the whole network is trained for 100 epochs with the Adam optimizer. The default initial learning rate is set to 0.01 and decayed by 0.1 for every 40 epochs. All the experiments are trained with the default batch size to 12 on 4 GPUs.

\paragraph{Dataset Setup} For the image hazing task, we find most previous state-of-the-art methods leverage available depth datasets to synthesize their own hazy datasets based on the physical corruption model in \Eref{eq:dehaze_model}, and conduct evaluation only on these specific datasets. Direct comparisons on these datasets are not fair. Recently, \cite{li2017reside} proposes a image dehazing benchmark RESIDE, which consists of large-scale training and testing hazy image pairs synthesized from depth and stereo datasets. To compare with state-of-the-art methods, they use many different evaluation metrics and conduct comprehensive comparisons among them. Although their test dataset consists of both indoor and outdoor images, they only report the quantitative results for the indoor parts. Following their strategy, we also compare our method on indoor dataset quantitatively and outdoor dataset qualitatively.

Similar to image hazing, there also exist several different large-scale synthetic datasets for image deraining. Most recently, \cite{zhang2018density} has developed a new dataset containing raining density labels (e.g. light, medium and heavy) for density-aware image deraining. Although we do not need the rain-density label information in our method, we still adopt this dataset for fair comparison. In this dataset, a total of 12000 training rainy images are synthesized with different orientations and scales with Photoshop.

\renewcommand{\arraystretch}{1.2}
\begin{table*}[t]
\begin{center}
\begin{tabular}{c|ccccccc}
\hline
& DCP \cite{he2011single} & CAP \cite{zhu2015fast} & GRM \cite{chen2016robust} & AOD-Net \cite{li2017aod} & DehazeNet \cite{cai2016dehazenet} & GFN \cite{ren2018gated} & \textbf{GCANet} \\
\hline
PSNR & 16.62 & 19.05 & 18.86 & 19.06 & 21.14 & 22.30 & \textbf{30.23}\\
\hline
SSIM & 0.82 & 0.84 & 0.86 & 0.85 & 0.86 & 0.88 & \textbf{0.98} \\
\hline
\end{tabular}
\vspace{1em}
\caption{Quantitative comparisons of image dehazing on the SOTS indoor dataset from RESIDE. Obviously, Our \textbf{GCANet} outperforms all the previous state-of-the-art image dehazing methods by a very large margin.}
\label{table:dehaze}
\end{center}
\end{table*}

\begin{figure*}[t]
\begin{center}

\includegraphics[width=1\linewidth]{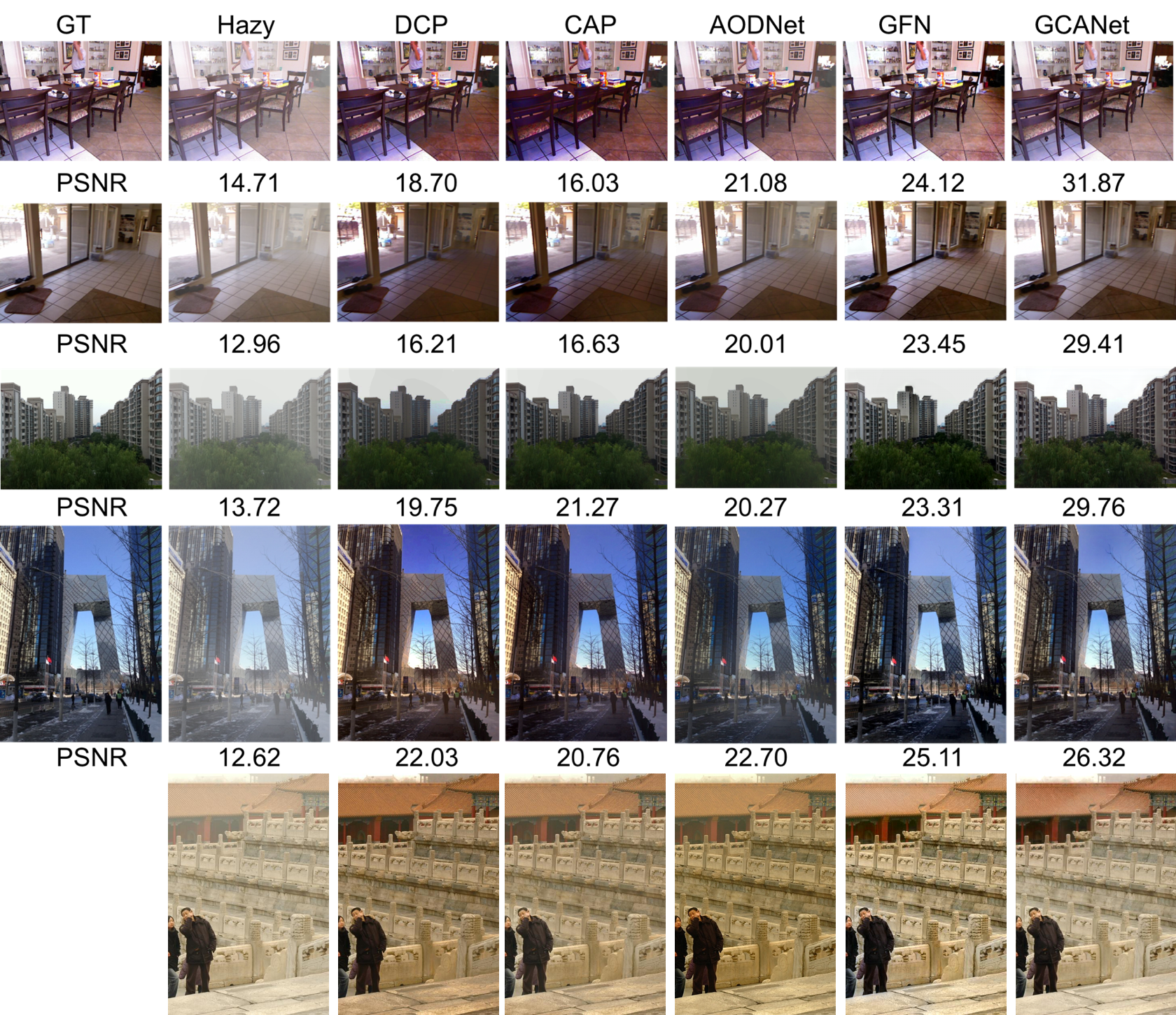}
\end{center}
\caption{Qualitative comparisons with different dehazing methods for indoor and outdoor hazy images, and the last row is one real hazy example. It can be seen that our \emph{GCANet} is the best one which can remove the underlying haze while maintaining the original brightness.}
\label{fg:dehaze}
\end{figure*}

\begin{table*}[ht]
\begin{center}
\begin{tabular}{cccccccc}
\hline
 DSC\cite{luo2015removing} & GMM \cite{li2016rain} & CNN\cite{fu2017clearing} & JORDER\cite{yang2017deep} & DDN \cite{fu2017removing}& JBO \cite{zhu2017joint} & DID-MDN\cite{zhang2018density} & \textbf{GCANet} \\
\hline
 21.44 & 22.75 & 22.07 & 24.32 & 27.33 & 23.05  & 27.95 & \textbf{31.68}\\
\hline
\end{tabular}
\vspace{1em}
\caption{Quantitative comparison results (PSNR) of the image deraining task on the DID-MDN test dataset. Although our \textbf{GCANet} is mainly designed for image dehazing, it generalizes very well for the image deraining task. }
\label{table:derain}
\end{center}
\end{table*}

\begin{figure*}[ht]
\begin{center}

\includegraphics[width=1\linewidth]{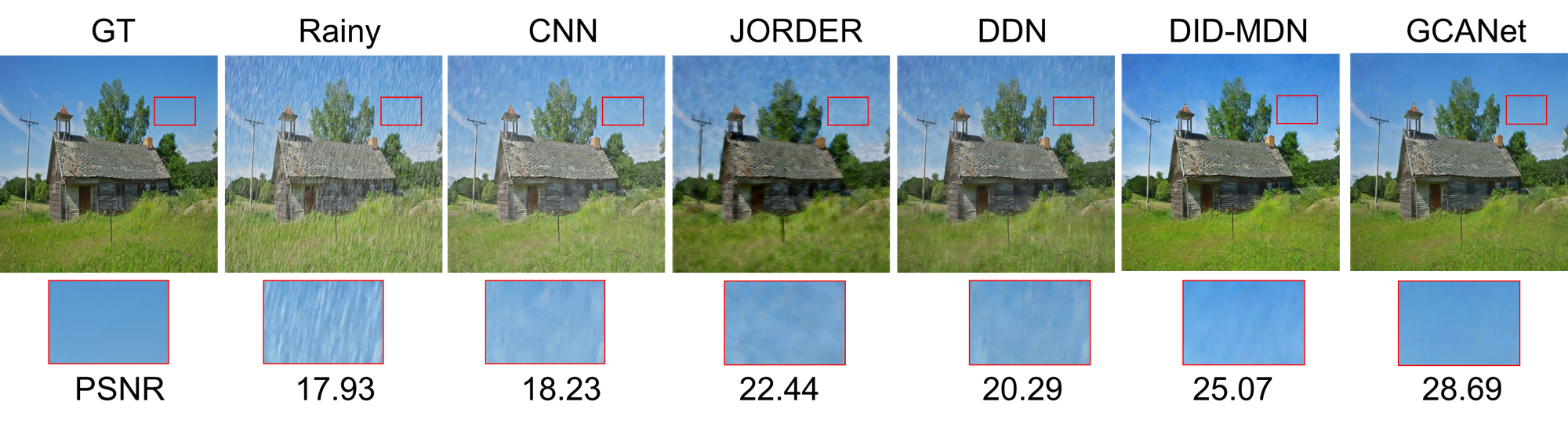}
\end{center}
\caption{One visual example deraining result for the different state-of-the-art deraining methods. Obviously, previous methods like CNN \cite{fu2017clearing}, JORDER \cite{yang2017deep} tend to under-derain the image, and our \textbf{GCANet} can achieve the best deraining results.}
\label{fg:derain}
\end{figure*}

\paragraph{Quantitative and Qualitative Evaluation for image dehazing} In this part, we will compare our method with previous state-of-the-art image dehazing methods both quantitatively and qualitatively. 

As shown in \Tref{table:dehaze}, six different state-of-the-art methods are used for quantitative evaluation: DCP\cite{he2011single}, CAP \cite{zhu2015fast}, GRM \cite{chen2016robust}, AOD-Net \cite{li2017aod}, DehazeNet \cite{cai2016dehazenet}, and GFN \cite{ren2018gated}. Among them, the first three are traditional prior-based methods and the last three are learning-based methods. For convenience, all the results except GFN shown in the \Tref{table:dehaze} are directly cited from \cite{li2017reside}. For GFN \cite{ren2018gated}, the latest state-of-the-art dehazing method, they have also reported the results on the RESIDE SOTS indoor dataset in their paper. Although various evaluation metrics are proposed in \cite{li2017reside}, we only adopt PSNR and SSIM, the most widely used metrics in previous methods. It can be seen that our proposed \textbf{GCANet} outperforms all previous dehazing methods by a large margin. 

We further show the dehazing results of two indoor and three outdoor hazy images in \Fref{fg:dehaze} for qualitative comparisons. From these visual results, we can easily observe that DCP \cite{he2011single} and CAP \cite{zhu2015fast} will make the  brightness of the dehazed results relatively dark, which is because of their underlying prior assumptions. For AOD-Net \cite{li2017aod}, we find that it is often unable to entirely remove the haze from the input. Although GFN \cite{ren2018gated} can achieve quite good dehazing results in some cases, our \textbf{GCANet} is the best one which can both preserve the original brightness and remove the haze as much as possible from the input.

\paragraph{Ablation Analysis} To understand the importance of each component in our \textbf{GCANet}, we have conducted ablation analysis with and without each specific component. Specifically, we focus on three major components: with / without the smoothed dilation, with / without the gated fusion sub-network, and with instance normalization / batch normalization. Correspondingly, four different network configurations are evaluated on the image dehazing task, and we incrementally add one component to each configuration at a time. As shown in \Tref{table:ablation}, the final performance keeps raising in these  experiments. However, one interesting observation is that it seems the biggest gain comes from instance normalization in place of batch normalization. Therefore, we further add one experiment by using instance normalization only without smoothed dilation and gated fusion network. Unsurprisingly, it can still achieve slightly better results than the first configuration with batch normalization, but the gain is smaller than the aforementioned one. That is to say, by combing all the designed components together, larger gains can be achieved than only applying one or some of them.

\setlength{\tabcolsep}{2.8pt}
\renewcommand{\arraystretch}{1.2}
\begin{table}[h]
\begin{center}
\begin{tabular}{cccccc}
\hline
smoothed dilation & &$\surd$&$\surd$& $\surd$  & \\
gated fusion      & &  &$\surd$& $\surd$  & \\
instance norm & &   &&  $\surd$&$\surd$ \\
\hline
PSNR & 27.57 & 28.12  & 28.72 & 30.23 & 28.45 \\
\end{tabular}
\caption{Detailed ablation analysis for each component with different training configurations, which shows that the combination of all the designed components is the best.}

\label{table:ablation}
\end{center}
\end{table}

\begin{figure}[h]
\begin{center}

\includegraphics[width=1\linewidth]{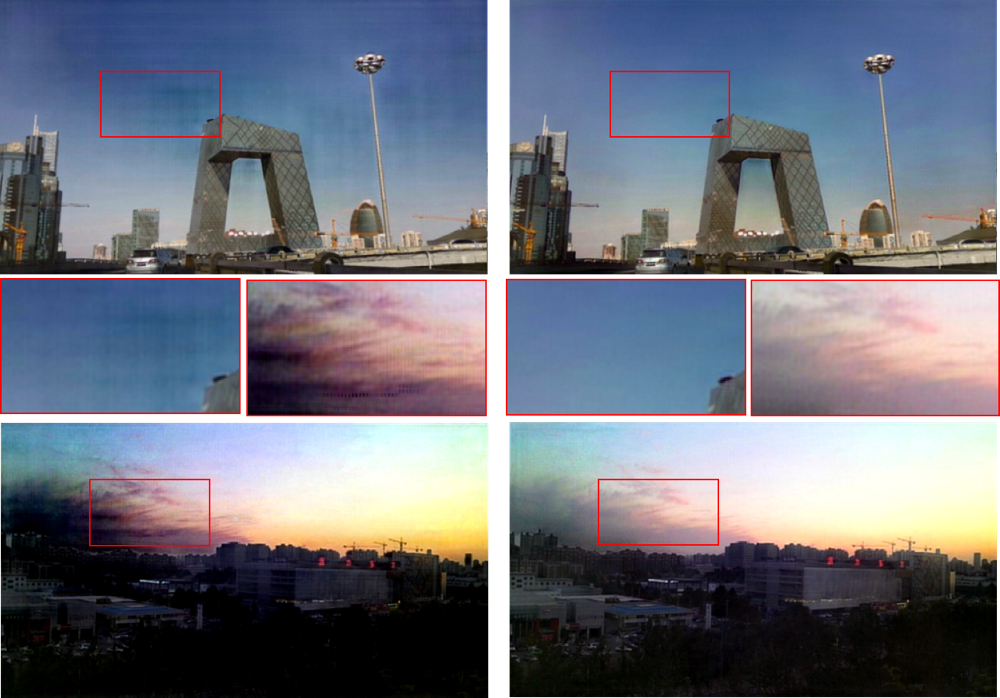}
\end{center}
\caption{Two dehazing examples to  show the superority of smoothed dilated resblocks (right column) and regular exponentially dilated resblocks (left colum). Obviously, our smoothed dilated resblocks improve the gridding artifacts and produce much better dehazing results.}
\label{fg:smooth_effect}
\end{figure}

To further validate the effectiveness of our smoothed dilated resblock in alleviating the gridding artifacts, we compare it with the previous widely-used exponentially dilated resblock \cite{chen2017fast,fan2018decouple,li2018recurrent}, where the dilation rates of adjacent resblocks are increased exponentially (e.g., 2, 4, 8, 16, 32). As shown in the two representative dehazing examples in \Fref{fg:smooth_effect}, the gridding artifacts and color shift often happen near the object boundaries and texture regions when the exponentially dilated resblocks are used. By contrast, our smoothed dilated resblocks can address this problem and preserve the original color fidelity.

\paragraph{Generality to Image Deraining Task} The task of image deraining is very similar to image dehazing, which aims to remove the rain-streak component from a corrupted image captured in the rainy environment. Though our focus is to design a good network structure for image dehazing, we are also very curious about whether the proposed \textbf{GCANet} can be applied to the image deraining task. Specifically, we leverage the training dataset synthesized in \cite{zhang2018density}, and compare our method with seven different image deraining methods: DSC \cite{luo2015removing}, GMM \cite{li2016rain}, CNN \cite{fu2017clearing}, JORDER \cite{yang2017deep}, DDN \cite{fu2017removing}, JBO \cite{zhu2017joint} and DID-MDN \cite{zhang2018density}. Note that all the results are cited from \cite{zhang2018density}. Surprisingly, as shown in \Tref{table:derain}, our \textbf{GCANet} even outperforms previous best method \cite{zhang2018density} with more than 3 dB in PSNR.

We also provide one deraining example in \Fref{table:derain} for visual comparison.  It can be seen that many previous methods like CNN \cite{fu2017clearing,fu2017removing} often tend to under-derain the image, and some unexpected patterns may appear in the deraining results of JORDER \cite{yang2017deep}. To see more details, we crop and zoom-in one local patch from the sky region. It is easy to observe that the deraining result of our \textbf{GCANet} is much clearer than other methods.
\section{Conclusion}
In this paper, we propose an end-to-end gated context aggregation network for image dehazing. To eliminate the gridding artifacts from the dilated convolution, a latest smoothed dilated technique is used. Moreover, a gated sub-network is leveraged to fuse the features of different levels. Despite of the simplicity of the proposed method, it is better than the previous state-of-the-art image dehazing methods by a large margin. We further apply the proposed network to the image deraining task, which can also obtain and state-of-the-art performance. In the future, we will try more facy losses used in \cite{chen2017stylebank,he2018deep} and consider to extend to video dehazing like \cite{chen2017coherent}. 
{\small
\bibliographystyle{ieee}
\bibliography{egbib}
}

\end{document}